\documentclass[letterpaper, 10 pt, conference]{ieeeconf}  

\IEEEoverridecommandlockouts                              

\newcommand{\Skip}[1]{}

\usepackage{graphicx}
\usepackage{amsmath}
\usepackage{amssymb}
\usepackage{booktabs}
\usepackage{bbding}
\usepackage{color,xcolor}
\usepackage{multirow}
\usepackage{float}
\usepackage{adjustbox}
\usepackage{url}
\usepackage{diagbox}
\usepackage[switch]{lineno}
\usepackage[breaklinks=true,colorlinks,linkcolor=black,citecolor=black,bookmarks=false]{hyperref}
\usepackage[capitalize]{cleveref}
\usepackage[style=ieee]{biblatex}
\AtNextBibliography{\footnotesize}

\crefname{section}{Sec.}{Secs.}
\Crefname{section}{Section}{Sections}
\Crefname{table}{Table}{Tables}
\crefname{table}{Tab.}{Tabs.}

\title{\LARGE \bf
ImmFusion: Robust mmWave-RGB Fusion for 3D Human Body Reconstruction in All Weather Conditions
}

\author{Anjun Chen$^{1}$, Xiangyu Wang$^{1}$, Kun Shi$^{1}$, Shaohao Zhu$^{1}$, Bin Fang$^{2}$, Yingfeng Chen$^{3}$, \\Jiming Chen$^{1}$, Yuchi Huo$^{4}$, and Qi Ye$^{1}$ 
\thanks{$^{1}$State Key Laboratory of Industrial Control Technology, Zhejiang University, Hangzhou, China.}%
\thanks{$^{2}$Tsinghua University, Beijing, China}%
\thanks{$^{3}$Fuxi AI Lab, NetEase, Hangzhou, China}%
\thanks{$^{4}$State Key Lab of CAD\&CG, Zhejiang University and Zhejiang Lab, Hangzhou, China}%
\thanks{Corresponding author: \emph{Qi Ye} (qi.ye@zju.edu.cn). Qi Ye is with the College of Control Science and Engineering, the State Key Laboratory of Industrial Control Technology, Zhejiang University, and also the Key Laboratory of Collaborative Sensing and Autonomous Unmanned Systems of Zhejiang Province. This work was supported in part by NSFC under Grants 62088101, 62233013, and 62103372. Project page: https://chen3110.github.io/ImmFusion/index.html}
}

\begin{document}

\newcommand{\reflabel}{dummy} 


\newcommand{\seclabel}[1]{\label{sec:\reflabel-#1}}
\newcommand{\secref}[2][\reflabel]{Section~\ref{sec:#1-#2}}
\newcommand{\Secref}[2][\reflabel]{Section~\ref{sec:#1-#2}}
\newcommand{\secrefs}[3][\reflabel]{Sections~\ref{sec:#1-#2} and~\ref{sec:#1-#3}}

\newcommand{\eqlabel}[1]{\label{eq:\reflabel-#1}}
\renewcommand{\eqref}[2][\reflabel]{(\ref{eq:#1-#2})}
\newcommand{\Eqref}[2][\reflabel]{(\ref{eq:#1-#2})}
\newcommand{\eqrefs}[3][\reflabel]{(\ref{eq:#1-#2}) and~(\ref{eq:#1-#3})}

\newcommand{\figlabel}[2][\reflabel]{\label{fig:#1-#2}}
\newcommand{\figref}[2][\reflabel]{Fig.~\ref{fig:#1-#2}}
\newcommand{\Figref}[2][\reflabel]{Fig.~\ref{fig:#1-#2}}
\newcommand{\figsref}[3][\reflabel]{Figs.~\ref{fig:#1-#2} and~\ref{fig:#1-#3}}
\newcommand{\Figsref}[3][\reflabel]{Figs.~\ref{fig:#1-#2} and~\ref{fig:#1-#3}}

\newcommand{\tablelabel}[2][\reflabel]{\label{table:#1-#2}}
\newcommand{\tableref}[2][\reflabel]{Table~\ref{table:#1-#2}}
\newcommand{\Tableref}[2][\reflabel]{Table~\ref{table:#1-#2}}
\newcommand{\etal}{et al.}
\newcommand{\eg}{e.g.}
\newcommand{\ie}{i.e. }
\newcommand{\etc}{etc. }

\def\bfmu{\mbox{\boldmath$\mu$}}
\def\bftau{\mbox{\boldmath$\tau$}}
\def\bftheta{\mbox{\boldmath$\theta$}}
\def\bfdelta{\mbox{\boldmath$\delta$}}
\def\bfphi{\mbox{\boldmath$\phi$}}
\def\bfpsi{\mbox{\boldmath$\psi$}}
\def\bfeta{\mbox{\boldmath$\eta$}}
\def\bfnabla{\mbox{\boldmath$\nabla$}}
\def\bfGamma{\mbox{\boldmath$\Gamma$}}

%
%


\newcommand{\R}{\mathbb{R}}

\maketitle
\thispagestyle{empty}
\pagestyle{empty}

\begin{abstract}

3D human reconstruction from RGB images achieves decent results in good weather conditions but degrades dramatically in rough weather. Complementary, mmWave radars have been employed to reconstruct 3D human joints and meshes in rough weather. However, combining RGB and mmWave signals for robust all-weather 3D human reconstruction is still an open challenge, given the sparse nature of mmWave and the vulnerability of RGB images. In this paper, we present ImmFusion, the first mmWave-RGB fusion solution to reconstruct 3D human bodies in all weather conditions robustly. Specifically, our ImmFusion consists of image and point backbones for token feature extraction and a Transformer module for token fusion. The image and point backbones refine global and local features from original data, and the Fusion Transformer Module aims for effective information fusion of two modalities by dynamically selecting informative tokens. Extensive experiments on a large-scale dataset, mmBody, captured in various environments demonstrate that ImmFusion can efficiently utilize the information of two modalities to achieve a robust 3D human body reconstruction in all weather conditions. In addition, our method's accuracy is significantly superior to that of state-of-the-art Transformer-based LiDAR-camera fusion methods. 

\end{abstract}


\section{Introduction}

3D human body reconstruction is widely used in many practical robotic applications \cite{munaro20143d,pereira2019reconstructing,liu2021graph} like human-robot interaction, pose estimation, and motion capture. Currently, the most popular approach is to reconstruct from RGB images, due to the progress of computer vision technologies. Nevertheless, the performance of reconstruction using RGB images under rough circumstances is still limited, lying that the perception capability of RGB cameras will rapidly deteriorate in poor illumination or inclement weather conditions. On the flip side, as regressing depth from a single image is inherently an ill-posed problem, the 3D reconstruction based on monocular cameras is fairly complicated. 

Recently, mmWave radar has gained increasing popularity in wireless sensing areas, like autonomous driving \cite{shi2021geometry,shi2021road}, human activity recognition \cite{meng2020gait,cheng2021person}, and UAV \cite{Axelsson_2021_CVPR}. Simultaneously, the mmWave radar has demonstrated great potential in human body reconstruction tasks for keeping unaffected by adverse environments. Therefore, its reconstruction results are competitive with or superior to that from RGB images in particular cases, as prior work \cite{chen2022mmbody} reveals. 

Despite the depth measurement and resistance to extreme weather conditions, reconstruction using mmWave signals suffers from sparsity and multi-path effect, which restricts its high performance in normal scenes. For example, the range and angle resolution of the device used in \cite{chen2022mmbody} are only 0.2m and 2 degrees, respectively. 
Introducing RGB signals into the mmWave system could be a possible solution. Fusing the two modalities to combine their strengths should be the key to realizing robust 3D human body reconstruction in all weather conditions. 

\Skip{
However, as these two types of data are represented in different modalities, the combination strategy requires a careful design. To tackle the discrepancy of heterogeneous geometrics, most existing works adopt point-to-image projection. Specifically, this paradigm simply fuses point clouds and RGB pixel values by element-wise addition or channel-wise concatenation. Nevertheless, this convention discards some beneficial information of intact point features (e.g., depth information), which cannot fully release the potential of two modalities. Besides, there remain challenges like sparse points, missing body parts, and inconsistent point clouds which would result in fetching few or even wrong image features in the field of mmWave-based human mesh reconstruction according to \cite{chen2022mmbody}. Additionally, the low quality of image features in harsh circumstances like poor illumination would severely degrade the performance of the model. 
}

However, combining multi-modal information is not trivial. Most existing LiDAR-camera fusion approaches adopt point-to-image projection to fuse point clouds and RGB pixel values by element-wise addition or channel-wise concatenation. Nevertheless, directly attaching the sparse, noisy, randomly missing, and temporally flicking mmWave point cloud to the RGB image would degrade the extracted features~\cite{chen2022mmbody}, especially in harsh circumstances like poor illumination. Therefore, this strategy is not suitable for mmWave-RGB fusion.
Alternatively, Transformer \cite{vaswani2017attention} opens up new avenues for the research on multi-modal fusion methods \cite{li2022deepfusion,wang2022multimodal,bai2022transfusion,wang2022bridged,zhang2022cat}. These Transformer fusion frameworks, however, focus on LiDAR-camera fusion-based object detection, which is inapplicable for the mmWave-RGB fusion-based human body reconstruction task. 

To address these issues, we present ImmFusion, the first fusion solution to combine the mmWave point clouds and RGB images to robustly reconstruct the 3D human body in all weather conditions. In view of the background noise caused by the multi-path effect, we extract features from point clusters rather than individual points. Besides, in order to address the spatial-temporal misalignment of heterogeneous modalities, we fuse the dense image features with the sparse radar point cluster features through a well-devised Fusion Transformer Module utilizing both local and global features. 
In addition, we conduct extensive experiments on the large-scale mmWave-human body dataset \cite{chen2022mmbody}, with 20 subjects captured in 7 scenes, including extreme weather conditions like fog, rain, and night. We evaluate the performance of our fusion model under different scenes, and it outperforms single-modality, point-level, and LiDAR-camera fusion methods in all weather environments. Our contributions can be summarized as follows:
\begin{itemize}

\item We propose ImmFusion, the first state-of-the-art mmWave-RGB fusion method for 3D human body reconstruction in all weather conditions including severe environments like rain, smoke, and poor illumination.

\item We employ a well-devised Fusion Transformer Module to effectively fuse the global and local features extracted from mmWave and RGB modalities, and we design an ingenious Modality Masking Module to strengthen the robustness of the model across all scenarios.

\item We evaluate the ImmFusion on the large-scale mmWave dataset mmBody \cite{chen2022mmbody} and demonstrate that ImmFusion outperforms other non-fusion or LiDAR-camera fusion models in all weather conditions.
\end{itemize}

\section{Related Works}

\subsection{Human Mesh Reconstruction from RGB Images}


3D human body reconstruction from RGB images can be broadly categorized into parametric \cite{loper2015smpl,pavlakos2019expressive} and non-parametric approaches \cite{kolotouros2019convolutional,lin2021end,lin2021mesh}. In the former, a body model, such as SMPL \cite{loper2015smpl} or SMPL-X \cite{pavlakos2019expressive}, represents the human body and regresses the input images. However, it is still challenging to estimate precise coefficients from a single image \cite{kocabas2020vibe, pavlakos2022human}. To improve the reconstruction, researchers utilize more visual information or dense relationship maps \cite{zhang2020learning}. Instead, non-parametric approaches directly regress the vertices from an image. Most pioneers choose Graph Convolutional Neural Network \cite{kolotouros2019convolutional} to model the local interactions between neighboring vertices with an adjacency matrix. Recently, METRO \cite{lin2021end} utilizes a Transformer encoder to jointly model vertex-vertex and vertex-joint interaction globally. Mesh Graphormer \cite{lin2021mesh} puts a step further by proposing a graph-convolution-reinforced transformer encoder and adding image grid features for joints and mesh vertices.



\subsection{mmWave-based Human Sensing}

Because mmWave sensors are capable of working in extreme conditions like rain, smoke, and occlusion, they have been widely utilized for sensing applications like human monitoring and tracking \cite{zhao2019mid}, detection and identification \cite{cheng2021person}, and behavior recognition \cite{meng2020gait}. Recently, Xue \etal ~\cite{xue2021mmmesh} proposed an accessible real-time human mesh reconstruction solution employing commercial portable mmWave devices. However, this work's datasets are not public, and the capability of the reconstruction from the mmWave signals in bad conditions is not studied. To fill the gap, Chen \etal ~\cite{chen2022mmbody} present a large-scale mmWave human body dataset with paired RGBD images in various environments, which paves the way for further research on combing mmWave radars with RGBD cameras for 3D body reconstruction.

\subsection{Fusion Methods for 3D Object Detection}
The bulk of literature is dedicated to mmWave-RGB fusion, which can be broadly divided into data, feature, and decision levels. 
1) The data-level fusion can be further categorized into radar-to-camera projection \cite{cheng2021robust,long2021radar} and camera-to-radar mapping \cite{zhang2021rvdet}. 
2) The decision-level fusion methods \cite{nabati2021centerfusion} usually leverage one modality to generate ROI containing valid objects, then make use of the other modality inside the ROI.  
3) The feature-level fusion derives from a pioneering framework dubbed AVOD \cite{ku2018joint}, where a series of anchors are predefined in the front-view map and the BEV map, respectively. 
Then, proposal-wise features are extracted within the region proposals and aggregated by specific fusion operations \cite{kim2020grif}.

Recently, the success of Transformer~\cite{vaswani2017attention} draws much attention to Transformer-based LiDAR-camera fusion. 
Specifically, DeepFusion \cite{li2022deepfusion} fuses deep camera and LiDAR features instead of decorating raw LiDAR points at the input level. Therein, 
\Skip{LearnableAlign is introduced that leverages the cross-attention mechanism to dynamically capture the correlations between two modalities, so as to better align the information from LiDAR features with the most related camera features.}
LearnableAlign is introduced that leverages the cross-attention mechanism to dynamically correlate LiDAR 
information with the most related camera features. 
Besides, TokenFusion \cite{wang2022multimodal} first prunes single-modal transformers, assuming that they can preserve information better, and further re-utilizes the pruned units for multimodal fusion. 
Moreover, TransFusion \cite{bai2022transfusion} conducts LiDAR-camera fusion with a soft-association approach to cope with inferior image situations.
These works, however, differ from ours since we make efforts in constructing a general mmWave-RGB fusion pipeline for 3D human body reconstruction.

\begin{figure*}
\centering
\setlength{\abovecaptionskip}{0.1mm}
\setlength{\belowcaptionskip}{-6mm}
    \includegraphics[width=0.9\linewidth]{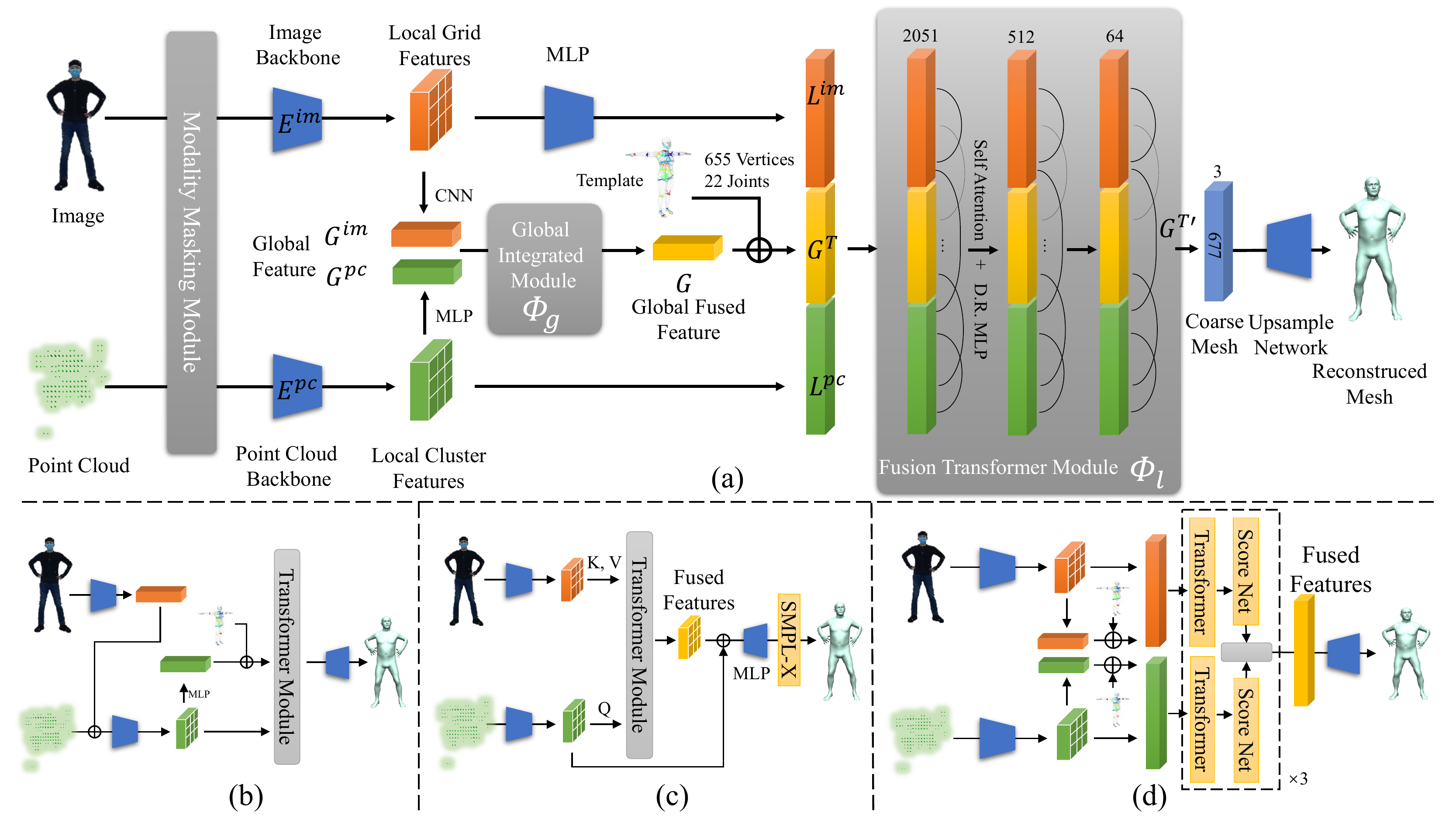}
    \caption{Comparison of different fusion strategies. (a) Our proposed ImmFusion. D.R. MLP stands for a dimension reduction MLP. (b) Point-level fusion methods \cite{wang2021pointaugmenting}. (c) DeepFusion \cite{li2022deepfusion}. (d) TokenFusion \cite{wang2022multimodal}.}
    \label{fig:framework}
\end{figure*}

\section{ImmFusion}

In this section, we present our proposed method ImmFusion for 3D human body reconstruction with both RGB images and mmWave point clouds as input. \cref{fig:framework} (a) illustrates the framework of ImmFusion. The structure aims to efficiently diffuse the image and point cloud features at global and local levels to predict the human body mesh. Given a radar point cloud with fixed numbers of points and an image with a size of $224\times224$, the global/local point and image features are firstly extracted by the image and point backbone, respectively. Next, the two global features are incorporated as one global feature vector and embedded with SMLP-X template positions. Then, all global/local features are tokenized as input of a multi-layer Fusion Transformer Module to dynamically fuse the information of two modalities and directly regress the coordinates of 3D human joints and coarse mesh vertices. Last, we employ Multi-Layer Perceptrons (MLPs) to upsample the coarse mesh vertices to the full SMPL-X \cite{pavlakos2019expressive} mesh vertices.

\subsection{Preliminary of 3D Human Body Reconstruction}

3D human body reconstruction aims to predict the 3D positions of all the joints and vertices. We adopt the non-parametric approach mentioned above for body reconstruction. As our focus is reconstruction, we use the bounding boxes automatically annotated from the ground-truth mesh joints to crop the region of interest containing only the body part. Given a dataset $\mathcal{D}=\left\{P_{t}, I_{t}, J_{t}, V_{t}\right\}, t=0, \ldots, N$, where $P_{t}\in \mathbb{R}^{1024 \times 3}, I_{t}\in \mathbb{R}^{224 \times 224 \times 3}$ are the cropped body region of the mmWave radar point cloud with 1024 points and the RGB image with a size of $224\times224$, and $J_{t} \in \mathbb{R}^{22 \times 3}, V_{t}\in \mathbb{R}^{10475 \times 3}$ are the annotation locations of 22 joints and 10475 vertices at time $t$, we make efforts to fuse information from two modalities of input to reconstruct 3D human body.

\subsection{Extraction of Global and Local Features}

Early point-level fusion works \cite{wang2021pointaugmenting, qi2020imvotenet} concatenate image features or projected RGB pixels to the point clouds as extended features of the point-based model, as \cref{fig:framework} (b) illustrates. However, this fusion strategy is not suitable for mmWave-RGB fusion due to the sparsity and noise of radar points. As discussed in \cite{chen2022mmbody}, undesirable issues like randomly missing and temporally flicking would lead to fetching fewer or even wrong image features. Additionally, the low quality of image features in adverse environments like poor lighting would severely degrade the performance of the model. Because of the imbalanced image feature qualities for different body parts, the reconstruction errors of different joints significantly vary. Instead, the point features extracted by networks like PointNet++ \cite{qi2017pointnet++} are relatively more balanced at different parts. Therefore, we propose fusing the image features and point features to improve the accuracy of the whole body.  

We extract global and local features for the image and point cloud inputs to help extract global contextual dependencies and model local interactions. Specifically, we directly feed point clouds and images to the commonly used point and image backbones to extract global and local features. Either backbone can be substituted with alternative options as necessary. For brevity, we leave out the subscript $t$ in the following parts.

For the point cloud data, we obtain cluster features $L^{pc} \in \mathbb{R}^{32\times(3+2048)}$, from a radar point cloud $P$ using PointNet++ $E^{pc}$, where 32 denotes the number of seed points sampled by Farthest Point Sample (FPS), 3 denotes the spatial coordinate, and 2048 denotes the dimension of features extracted from the grouping local points. 
A global feature vector $G^{pc} \in \mathbb{R}^{2048}$ is further extracted from cluster features $L^{pc}$ using an MLP. 
For image data, we acquire the global feature $G^{im} \in \mathbb{R}^{2048}$ and grid features $L^{im} \in \mathbb{R}^{49\times 2051}$ using HRNet \cite{wang2020deep}  $E^{im}$. (MLPs are used to make features from HRNet the same as that of the point features.)

The two global features are fused into a global feature $G \in \mathbb{R}^{2048}$ by Global Integrated Module (GIM) $\Phi_g$ implemented using a tiny Transformer module,
\begin{equation}
 G =\Phi_g (G^{im}, G^{pc} ).
\end{equation}
Following \cite{kolotouros2019convolutional}, we perform positional encoding by attaching 3D coordinates of each joint and vertex in a human template mesh to the global vector $G^{T} = cat(J^{template}, V^{template}, G)$, where $G^{T} \in \mathbb{R}^{677 \times 2051}$. Both local features serve the purpose of providing fine-grained local details for body reconstruction.

\begin{figure*}
\centering
\setlength{\abovecaptionskip}{0.1mm}
\setlength{\belowcaptionskip}{-6mm}
    \includegraphics[width=0.9\linewidth]{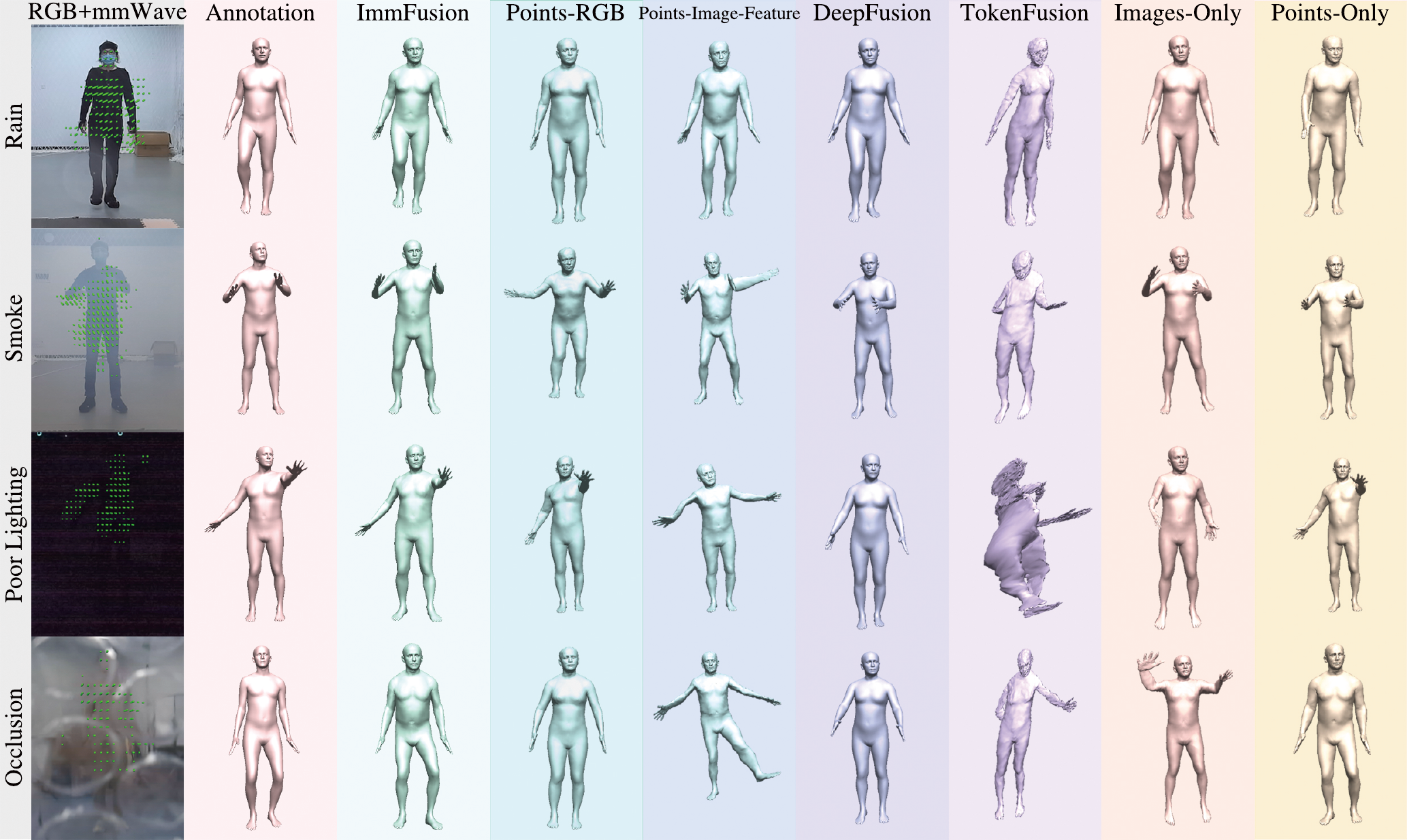}
    \caption{Qualitative results. Each row represents an adverse weather scene(rain, smoke, poor lighting, and occlusion) and each column shows the reconstructed mesh of the corresponding model, respectively.}
    \label{fig:goodcase}
\end{figure*}

\subsection{Transformer Fusion with Global and Local Features}

Multi-head attention module \cite{vaswani2017attention} is famous for modeling the relationship between information tokens. We adopt this structure to mitigate the feature degradation caused by the sparsity and noise of mmWave signals and the deficiency of RGB information in extreme conditions. We utilize the Fusion Transformer Module $\Phi_l$ to combine the strengths of radar points and images, enabling the model to select informative token features from two modalities dynamically:
\begin{equation}
G^{T^{\prime}}, L^{im\prime}, L^{pc\prime}=\Phi_{l}\left(G^{T}, L^{im}, L^{pc}\right),
\end{equation}
where $G^{T^{\prime}}\in \mathbb{R}^{677 \times 64}$, $L^{im\prime}\in \mathbb{R}^{49 \times 64}$ and $L^{pc\prime} \in \mathbb{R}^{32 \times 64}$. While attending to valid features and restricting undesirable features, the Fusion Transformer Module $\Phi_l$ adaptively adopts cross attention between joint/vertex queries $G^{T}$ generated from global features $G$ and point/image token features from local features $L^{im}$ $L^{pc}$  to aggregate relevant contextual information. Simultaneously, the self-attention mechanism reasons interrelations between each pair of candidate queries. 
Then, we adopt a dimension-reduction architecture, Graph Convolution \cite{kolotouros2019convolutional}, to decode the queries $G^{T^{\prime}}$ containing rich cross-modalities information into 3D coordinates of joints and vertices. 
Last, a linear projection network implemented using MLPs upsamples the coarse output mesh to the original 10475 vertices.

\subsection{Distortion Solution by Modality Masking}

Despite the superiority of the multi-head attention mechanism, the model is prone to struggle with sensor distortions according to \cite{bijelic2020seeing} due to the bias of training data (without data under adverse conditions), which makes Transformer focus all attention on the single modality that performs better under normal circumstances as demonstrated in our experiments. To effectively activate the model's adaptability across general scenarios, we design a Modality Masking Module (MMM) to mask one of the input modalities randomly and thus enforce the model to learn from the other modality in various situations. As a result, MMM enables the Fusion Transformer Module to overcome the training data bias problem and consider both modalities, which further facilitates the model to perform better across all scenarios in our experiments. In addition to the modality masking, we also randomly mask some percentages of input token features to simulate self or smoke occlusions and missing parts.

\subsection{Comparison with Relevant Methods}
We compare our proposed ImmFusion with other relevant fusion methods. Point-level fusion methods are implemented by decorating point clouds with RGB values fetched by projecting 3D point clouds to the original image plane \cite{wang2021pointaugmenting} or image features extracted by the 2D CNN backbone and feeding them directly to the Points-Only pipeline as \cref{fig:framework} (b) shows. 
We also compare ImmFusion with the LiDAR-camera fusion methods, \ie DeepFusion \cite{li2022deepfusion} and TokenFusion \cite{wang2022multimodal}, which show the state-of-the-art accuracy in the task of 3D object detection. As DeepFusion is a generic Transformer-based block that is incompatible with the dimension-reduction mechanism, we adopt the parametric reconstruction pipeline by replacing the detection framework of DeepFusion with linear projection to regress SMPL-X parameters. The framework of DeepFusion for 3D body reconstruction is shown in \cref{fig:framework} (c).
As for the TokenFusion method, we mainly implement it by referring to the scoring function in \cite{wang2022multimodal}. Specifically, we plug the scoring net among the Transformer layers of single-modality models to dynamically predict the importance of joint/vertex tokens and substitute inferior tokens with corresponding ones from the other modality as \cref{fig:framework} (d) suggests.


\begin{table*}[ht]
\setlength{\abovecaptionskip}{0.1mm}
  \centering
    \caption{Errors (cm) of different methods for 3D body reconstruction in different scenes. For the two columns of each scene, the first column is for joint error and the second vertex error.}
  \resizebox{\textwidth}{!}{
    \begin{tabular}{c|c|cc|cc|cc|cc|cc|cc|cc|cc}
    \toprule
    \multicolumn{2}{c|}{\multirow{2}[2]{*}{Scenes}} & \multicolumn{6}{c|}{Basic Scenes}             & \multicolumn{8}{c|}{Adverse Environments}                     & \multicolumn{2}{c}{\multirow{2}[2]{*}{Average}} \\
    \multicolumn{2}{c|}{} & \multicolumn{2}{c}{ \textbf{Lab1}} & \multicolumn{2}{c}{ \textbf{Lab2}} & \multicolumn{2}{c|}{ \textbf{Furnished}} & \multicolumn{2}{c}{ \textbf{Rain}} & \multicolumn{2}{c}{ \textbf{Smoke}} & \multicolumn{2}{c}{ \textbf{Poor lighting}} & \multicolumn{2}{c|}{ \textbf{Occlusion}} & \multicolumn{2}{c}{} \\
    \midrule
    \multirow{10}[4]{*}{Mean Error} & Points-RGB & 6.7   & 9.3   & 6.7   & 8.7   & 6.6   & 8.9   & 7.7   & 10.1  & 11.3  & 14.8  & 7.0   & 9.4   & 12.0  & 17.2  & 8.3   & 11.2 \\
          & Points-Image-Feature & 4.4   & 6.1   & 4.2   & 5.4   & 6.0   & 8.0   & 6.4   & 8.5   & 8.0   & 10.9  & 13.0  & 19.6  & 18.4  & 20.7  & 8.6   & 11.3 \\
          & DeepFusion\cite{wang2020deep} & 5.1   & 6.5   & 5.7   & 6.8   & 6.7   & 8.2   & 7.0   & 8.2   & 9.6   & 12.1  & 13.4  & 16.9  & 13.3  & 17.8  & 8.7   & 10.9 \\
          & TokenFusion\cite{wang2022multimodal} & 4.3   & 6.0   & 4.0   & 5.3   & 5.6   & 7.0   & 6.0   & 7.4   & 9.4   & 12.9  & 11.3  & 15.7  & 10.8  & 14.9  & 7.4   & 9.9 \\
          & Images-Only & 4.1   & 5.5   & 4.0   & 5.3   & 5.4   & 6.8   & 5.9   & 7.4   & 8.5   & 11.2  & 9.9   & 14.1  & 11.3  & 16.6  & 7.0   & 9.6 \\
          & Points-Only & 6.3   & 8.8   & 6.7   & 8.9   & 6.4   & 8.8   & 7.8   & 10.2  & 8.0   & 10.6  &  \textbf{6.2} &  \textbf{8.4} & 8.8   & 12.7  & 7.2   & 9.8 \\
\cmidrule{2-18}          & ImmFusion-w/o-LF & 4.9   & 6.5   & 4.7   & 6.0   & 6.0   & 7.8   & 6.7   & 8.1   & 8.5   & 10.9  & 10.9  & 15.5  & 10.4  & 14.4  & 7.4   & 9.9 \\
          & ImmFusion-w/o-MMM & 4.1   & 5.7   & 3.8   & 5.0   & 5.3   & 7.0   & 6.0   & 7.2   & 7.9   & 10.1  & 9.7   & 13.6  & 10.7  & 14.1  & 6.8   & 9.0 \\
          & ImmFusion-w/o-GIM & 4.1   & 5.5   & 3.7   & 4.8   & 5.3   & 6.6   & 6.1   & 7.3   & 7.7   &  \textbf{9.7} & 7.6   & 9.5   & 9.6   & 14.9  & 6.3   & 8.3 \\
          & ImmFusion &  \textbf{4.1} &  \textbf{5.4} &  \textbf{3.7} &  \textbf{4.7} &  \textbf{5.2} &  \textbf{6.4} &  \textbf{5.6} &  \textbf{6.8} &  \textbf{7.6} & 9.8   & 6.8   & 9.0   &  \textbf{7.8} &  \textbf{11.0} &  \textbf{5.9} &  \textbf{7.4} \\
    \midrule
    \multirow{10}[4]{*}{Max Error} & Points-RGB & 24.8  & 34.3  & 27.0  & 38.7  & 22.6  & 31.5  & 29.2  & 38.2  & 38.8  & 54.7  & 24.4  & 33.2  & 36.4  & 47.9  & 29.0  & 39.8 \\
          & Points-Image-Feature & 13.2  & 19.2  & 13.1  & 19.1  & 18.2  & 25.7  & 22.2  & 29.1  & 22.9  & 32.6  & 60.2  & 82.8  & 72.5  & 79.6  & 31.8  & 41.1 \\
          & DeepFusion\cite{wang2020deep} & 10.8  & 15.6  & 12.9  & 18.4  & 13.9  & 19.9  & 22.0  & 27.9  & 20.2  & 28.2  & 37.8  & 55.2  & 41.1  & 57.1  & 22.7  & 31.7 \\
          & TokenFusion\cite{wang2022multimodal} & 12.7  & 17.8  & 13.5  & 19.9  & 15.0  & 21.6  & 21.7  & 26.9  & 26.5  & 36.9  & 46.3  & 69.4  & 38.5  & 59.4  & 24.9  & 36.0 \\
          & Images-Only & 11.3  & 15.5  & 12.1  & 17.2  & 14.0  & 18.9  & 21.7  & 27.5  & 23.2  & 31.2  & 43.2  & 64.5  & 48.6  & 73.8  & 24.9  & 35.5 \\
          & Points-Only & 22.8  & 32.5  & 28.2  & 41.8  & 22.3  & 31.5  & 29.4  & 39.1  & 25.7  & 36.6  &  \textbf{21.7} &  \textbf{30.3} & 28.3  & 38.4  & 25.5  & 35.8 \\
\cmidrule{2-18}          & ImmFusion-w/o-LF & 14.9  & 21.5  & 16.3  & 23.7  & 19.1  & 26.1  & 23.4  & 29.7  & 24.1  & 31.9  & 45.3  & 70.2  & 37.5  & 56.0  & 25.8  & 37.0 \\
          & ImmFusion-w/o-MMM &  \textbf{10.8}  & 16.2  & 10.6  & 16.0  & 13.7  & 20.0  & 21.4  & 26.7  & 21.2  & 29.7  & 39.6  & 56.5  & 39.3  & 57.3  & 22.4  & 31.8 \\
          & ImmFusion-w/o-GIM & 10.9  & 15.3  & 10.7  & 14.9  & 14.1  & 18.3  & 21.3  & 26.2  &  \textbf{20.2} &  \textbf{27.0} & 25.5  & 33.8  & 30.5  & 40.4  & 19.0  & 25.1 \\
          & ImmFusion & 10.9 &  \textbf{14.7} &  \textbf{10.5} &  \textbf{14.2} &  \textbf{13.6} &  \textbf{18.2} &  \textbf{20.1} &  \textbf{24.6} & 20.3  & 27.2  & 23.1  & 31.3  &  \textbf{27.0} &  \textbf{35.8} &  \textbf{17.9} &  \textbf{23.7} \\
    \bottomrule
    \end{tabular}%
    }
  \label{tab:errors}%
\end{table*}%

\section{Experiments}
\subsection{Experimental Settings}

\noindent \textbf{Dataset.} We conduct the experiments on the large-scale mmWave 3D human body dataset mmBody \cite{chen2022mmbody}, which consists of a considerable number of synchronized and calibrated mmWave radar point clouds and RGBD images in various conditions and mesh annotations for humans in the scenes. More specifications about the mmWave radar are provided in the product overview \cite{radar-overview}. 
We choose 10 sequences in the lab scenes as the training set while 2 sequences for each scene including labs, furnished, rain, smoke, poor lighting, and occlusion as the test set. 

\noindent \textbf{Metrics.} To evaluate the performance of the reconstruction, we employ the metrics of mean (max) joint error per frame and mean (max) vertex error per frame, which quantify the average (maximum) Euclidean distance between the prediction and the ground truth for joints/vertices in each frame. In comparison to the mean joint/vertex error, the max joint/vertex error is stricter.

\noindent \textbf{Implementation Details.}
Our ImmFusion applies $L_1$ loss to the reconstructed mesh to constrain the vertices and joints. In addition, the coarse meshes of each layer of the Fusion Transformer Module are also supervised by downsampled ground truth meshes using  $L_1$ loss to accelerate convergence.
All the models are implemented using Pytorch and are trained on an Nvidia GeForce RTX 3090. To be fair, we train all the networks for 50 epochs from scratch with an Adam optimizer and an initial learning rate of 0.001.

\subsection{Experimental Results}

\cref{fig:goodcase} shows the reconstructed meshes from ImmFusion for different poses and subjects in the different scenarios. Overall, the reconstructed meshes for most samples are close to the ground truth. \cref{tab:errors} summarizes the main results of all models tested on the mmBody dataset. Compared with existing fusion solutions and baselines, our approach can better exploit the complementary nature of two modalities: in addition to eliminating the negative effects of one modality on the other one, it also enhances the performance of one single modality by utilizing the complementary feature of the other. We provide more qualitative comparisons by visualizing the attention interactions between different token features in our accompanying video.

\noindent \textbf{Comparison with Single-modality Methods.} To demonstrate the effectiveness of our proposed fusion method, we compare ImmFusion with approaches using single-modality input. We implement single-modality methods by removing one input stream from our proposed ImmFusion pipeline. For the Images-Only input, the feature extractor consists of only the CNN backbone to extract image features. Regarding the Points-Only input, we substitute the CNN backbone with PointNet++. Experimental results demonstrate that ImmFusion is able to integrate the two modalities effectively. As shown in \cref{tab:errors}, the average of mean joint errors and mean vertex errors for ImmFusion can reach as low as 5.8cm and 7.6cm, decreasing by more than about 1cm and 2cm from that for Images-Only or Points-Only methods. Furthermore, ImmFusion achieves better accuracy of max errors than the other two types of single-modality methods across all scenes, illustrating that ImmFusion can dynamically select preferable information from mmWave point cloud and RGB images. Particularly for poor lighting and occlusion scenes where the RGB camera fails, ImmFusion can work robustly as the mmWave radar uses active lighting of mmWave frequency. 

\noindent \textbf{Comparison with Point-level Fusion Methods.} We compare ImmFusion with point-level fusion methods Points-RGB and Points-Image-Feature implemented by decorating point clouds with RGB values and image features. For the Points-RGB method, this intuitive fusion gains little accuracy improvement in basic scenes and even performs worse than the single-modality methods in adverse scenes, which is mainly due to the under-exploration of the inter-modality interaction. The Points-Image-Feature baseline makes a further step to integrate the multi-modal information, which gains accuracy in the basic scenes to some extent. However, the inferior image features in the severe scenes cannot be restricted by the attention module in this way, which severely degrades the performance.

\noindent \textbf{Comparison with LiDAR-camera Fusion Methods.} We also compare ImmFusion with the state-of-the-art fusion methods DeepFusion and TokenFusion. In spite of the simple fusion strategy, ImmFusion achieves more preferable results in all scenarios. DeepFusion tends to lose more global interactions due to the lack of global features. While TokenFusion aims to discard unimportant token features among Transformer layers, it is ineffective to incorporate the single-modality streams at the end of the model, which ultimately leads to unfavorable results as shown in \cref{fig:goodcase}.

\subsection{Ablation Study}
We conduct a comprehensive study to validate the effectiveness of local features, Modality Masking Module (MMM), and Global Integrated Module (GIM). 

\noindent \textbf{Effectiveness of Local Features.} 
The local features, which directly affect the quality and details, play a very important role in reconstruction tasks. To analyze the effectiveness of the local features, we compared the results of the original ImmFusion with its variation ImmFusion-w/o-LF, in which the cluster features and grid features are removed from the backward computation graph. As indicated in \cref{tab:errors}, the mean and max errors of ImmFusion-w/o-LF are obviously greater than ImmFusion. Despite the assistance of MMM, the errors in extreme conditions like poor lighting or occlusion are even worse than that of the Images-Only model. The max vertex error in poor lighting is up to 70cm, double higher than that of the original ImmFusion. These results strongly support our motivation of utilizing local features to benefit the quality of reconstruction.

\noindent \textbf{Effectiveness of Modality Masking Module.} 
An important question is whether MMM is valid. The results of single-modality methods, \ie Images-Only and Points-Only in \cref{tab:errors} report that RGB images have better accuracy than mmWave point clouds in the basic scenes due to the high resolution. Therefore, the training set only consisting of basic data would force the Transformer module to pay more attention to the image modality, which leads to a rapid decline of the performance in the poor lighting and occlusion scenes. Clearly, MMM eliminates the bias of training data and significantly improves the performance in extreme scenes as the result of ImmFusion-w/o-MMM demonstrates. Surprisingly, MMM gains the accuracy of the model across all scenes, which is mainly due to the fact that MMM enforces the Transformer module to lean more attention on the point modality to select helpful features. We train several models with varying maximum masking percentages to choose the best one and the optimal proportion is 30\%.

\begin{figure}
\centering
    \includegraphics[width=0.88\linewidth]{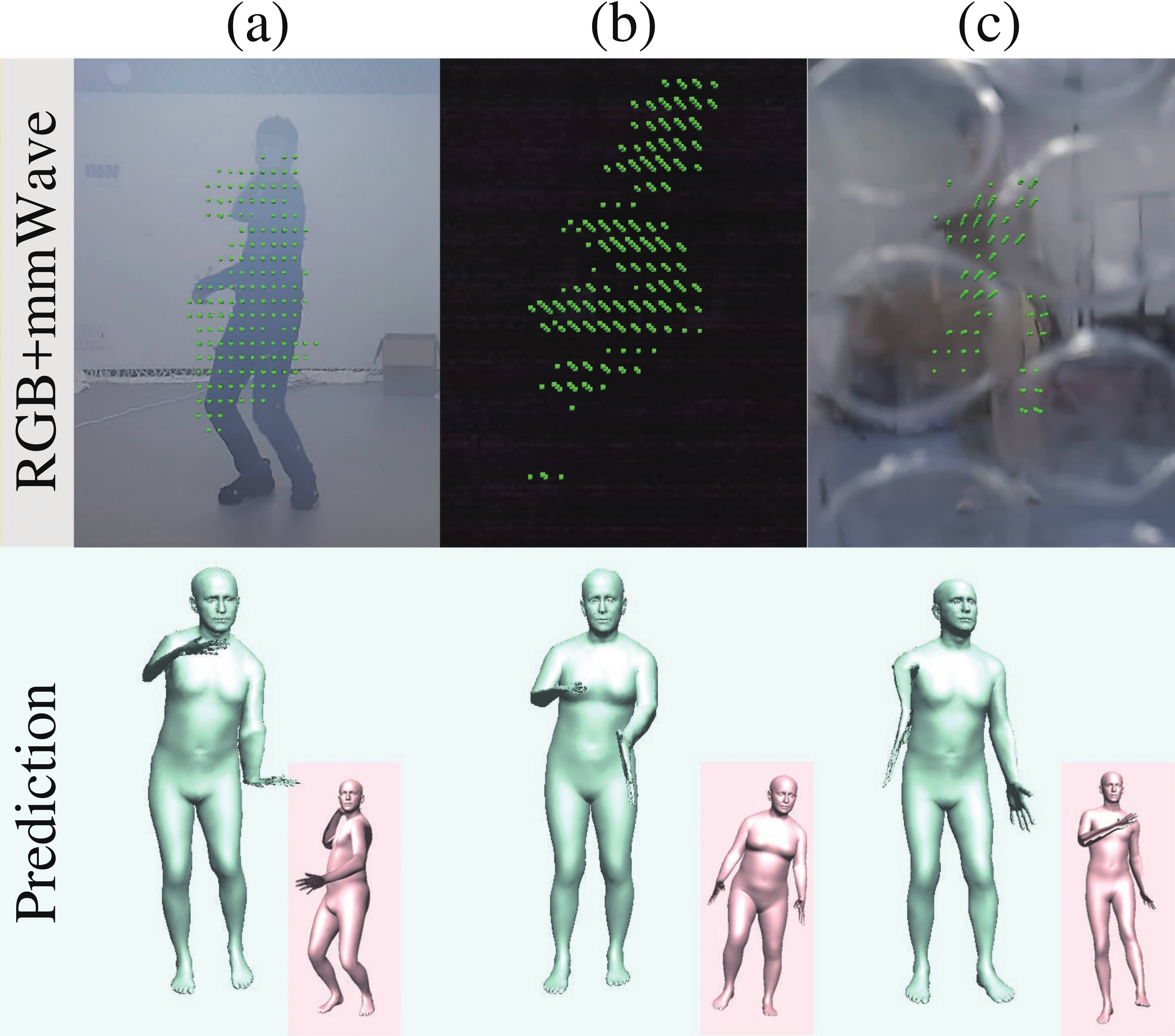}
    \caption{Failure cases. Columns (a) (b), and (c) show the unfavorable reconstruction results in the smoke, poor lighting, and occlusion scene respectively (Ground truth in pink).}
    \label{fig:badcase}
\end{figure}

\noindent \textbf{Effectiveness of Global Integrated Module.} 
In ImmFusion, GIM serves as a mixer to integrate global features of mmWave and RGB input. Instead of naive element-wise addition or channel-wise concatenation, GIM contains learnable parameters to control the weights of global features from different modalities. Among all types of scenes in \cref{tab:errors}, ImmFusion-w/o-GIM merely outperforms ImmFusion a little in the smoke environment, where the valid information proportion of RGB v.s mmWave is balanced, misleading the model to select useless features from the global feature. In other situations, especially in the poor lighting and occlusion scene, ImmFusion-w/o-GIM clearly underperforms compared with ImmFusion, proving the importance of GIM.

\subsection{Future Work}
There are situations when the reconstruction from ImmFusion fails, some of which are exampled in \cref{fig:badcase}. The reasons for these failures can be mainly attributed to the drawbacks of sensors. The low quality of images in rough conditions and the sparsity of mmWave point clouds severely complicate the regression task since non-parametric methods are prone to give non-smooth meshes when facing these challenges. Adding the third modality like depth images or LiDAR point clouds and exploiting a unified sensor fusion framework may settle this problem. Additionally, constrained by the data collection, we leave the extension of our method to the outdoor scenario as future work.

\section{CONCLUSIONS}

In this paper, we introduce ImmFusion, a multi-modal fusion model which combines mmWave and RGB signals for robust all-weather 3D human body reconstruction. In addition to the good results in basic scenes, ImmFusion shows great robustness in severe environments like rain, smoke, poor lighting, and occlusion due to the effectiveness of the attention mechanism and the Modality Masking Module. Experimental results suggest that ImmFusion can efficiently fuse the information of mmWave and RGB signals. In addition, we investigate various fusion approaches and demonstrate that ImmFusion outperforms single-modality, point-level, and LiDAR-camera fusion methods in all basic scenes and the majority of adverse environments. 






\printbibliography


\end{document}